%% file: main.tex
\documentclass[10pt,twocolumn,letterpaper]{article}

\usepackage{cvpr}              

\input{preamble}

\definecolor{cvprblue}{rgb}{0.21,0.49,0.74}
\usepackage[pagebackref,breaklinks,colorlinks,allcolors=cvprblue]{hyperref}
\usepackage{comment}
\usepackage{arydshln}
\usepackage{bbm}
\usepackage{amsthm}
\usepackage{multirow}
\usepackage{dsfont}
\usepackage[utf8]{inputenc}
\usepackage{csquotes}

\usepackage{verbatim}
\usepackage{algorithm}
\usepackage{algorithmic}
\newcommand{\minimize}[2]{\ensuremath{\underset{\substack{{#1}}}%
{\mathrm{minimize}}\;\;#2 }}
\newcommand\bs[1]{\boldsymbol{#1}}
 
\newcommand{\support}{\mathbb{S}}
\newcommand{\query}{\mathbb{Q}}
\newcommand{\p}{\mathrm{p}}	
\def\paperID{16744} 
\def\confName{CVPR}
\def\confYear{2025}

\title{UNEM: UNrolled Generalized EM for Transductive Few-Shot Learning}

\author[1,3]{Long Zhou}
\author[2]{Fereshteh Shakeri}
\author[3]{Aymen Sadraoui}

\author[3,4]{\authorcr Mounir Kaaniche}
\author[3]{Jean-Christophe Pesquet}
\author[2]{Ismail Ben Ayed}

\affil[1]{Politecnico di Milano}
\affil[2]{ÉTS Montréal}
\affil[3]{Université Paris-Saclay, Inria, CentraleSupélec, CVN}
\affil[4]{Université Sorbonne Paris Nord, L2TI, UR 3043}

\begin{document}
\maketitle
\input{sec/0_abstract}    
\input{sec/1_intro}

\input{sec/2_related_work}

\input{sec/3_methodology}
\input{sec/4_experiments}

\input{sec/5_conclusion}

{
    \small
    \bibliographystyle{ieeenat_fullname}
    \bibliography{main}
}

\input{sec/X_suppl}

\end{document}

%% file: preamble.tex
\usepackage{multirow}
\usepackage{graphicx}
\usepackage{booktabs}
\usepackage{multirow}
\usepackage{xcolor}
\usepackage{colortbl}
\makeatletter
\robustify\@latex@warning@no@line
\makeatother
\usepackage{authblk}

\usepackage[accsupp]{axessibility}  

%% file: sec/0_abstract.tex
\begin{abstract}


Transductive few-shot learning has recently triggered wide attention in computer vision. Yet, current methods introduce key hyper-parameters, which control the prediction statistics of the test batches, such as the level of class balance, affecting performances significantly. Such hyper-parameters are empirically grid-searched over validation data, and their configurations may vary substantially with the target dataset and pre-training model, making such empirical searches both sub-optimal and computationally intractable. In this work, we advocate and introduce the {\em unrolling} paradigm, also referred to as ``learning to optimize", in the context of few-shot learning, thereby learning efficiently and effectively a set of optimized hyper-parameters. Specifically, we unroll a generalization of the ubiquitous Expectation-Maximization (EM) optimizer into a neural network architecture, mapping each of its iterates to a layer and learning a set of key hyper-parameters over validation data. Our unrolling approach covers various statistical feature distributions and pre-training paradigms, including recent foundational vision-language models and standard vision-only classifiers. We report comprehensive experiments, which cover a breadth of fine-grained downstream image classification tasks, showing significant gains brought by the proposed unrolled EM algorithm over iterative variants. The achieved improvements reach up to 10\% and 7.5\% on vision-only and vision-language benchmarks, respectively. The source code and learned parameters are available at \url{https://github.com/ZhouLong0/UNEM-Transductive}.    

\end{abstract}

%% file: sec/1_intro.tex
\section{Introduction}
\label{sec:intro}


\begin{figure}[!h]
    \centering
    \includegraphics[width=0.45\textwidth]{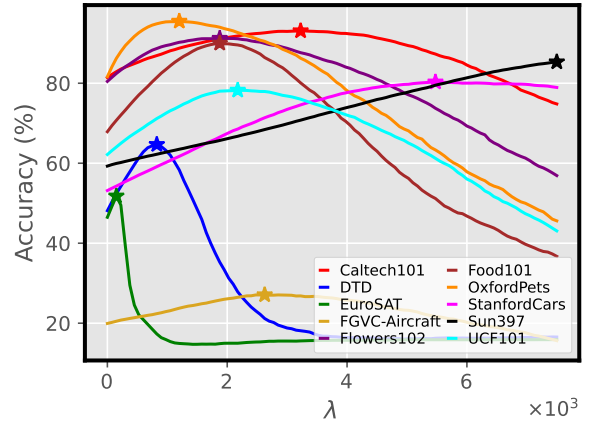} \vspace{-0.50cm}
    \caption{Impact of the class-balance hyperparameter $\lambda$ on the accuracy of transductive few-shot classification. The accuracy results are obtained using the EM-Dirichlet algorithm \cite{EMDIRICHLET_CVPR2024} applied to vision-language models (with 4-shots). The plot shows that the choice of $\lambda$ has a strong impact on the performance, and that the optimal $\lambda$ (indicated with the star symbol) might vary by orders of magnitudes, depending on the target downstream dataset (e.g., the ten very different fine-grained classification datasets). Further comments on the optimal $\lambda$ values and the values chosen in \cite{EMDIRICHLET_CVPR2024} are provided in Section \ref{sec:experiments}. The values of the learned hyper-parameters based on the proposed unrolled algorithm are illustrated and analyzed in Appendix \ref{app:add_res}.}
    \label{fig:acc_lambda}
\end{figure}

Deep learning has transfigured computer vision, driving substantial progress in tasks such as image classification, captioning, object detection and segmentation. However, these successes often come with a high cost: the requirement for large amounts of labeled data. Additionally, the generalization of these models is seriously challenged when evaluated on new classes (concepts), unseen during pre-training, or when operating under distribution shifts \cite{BAEK2022, WANGAI2024}.

To address these challenges, Few-Shot Learning (FSL) has recently attracted wide attention within the computer vision community. In the standard few-shot image classification setting, a feature extractor is first pre-trained on a set of classes, often called the base classes. Then, the model is adapted and evaluated on new classes and tasks. Each evaluation task is performed on a set of unlabeled samples (referred to as the query set), and supervised by a support set composed of few labeled samples per new class. Earlier FSL methods have relied on various concepts, including meta-learning \cite{Ravi2017OptimizationAA, MAML_ICML2017, JAMALTAML2019}, transfer learning \cite{Hariharan2017LowShotVR, 
MOZER2028TL, HangQi2018}, and metric learning \cite{mini-imagenet, FERIS2021MeL, XUEDML2023}. However, most of these works operate within the \textit{inductive} setting, where, at inference time, the prediction for each sample is made independently from the other samples in the query set. 

Recently, significant attention has shifted to the {\em transductive} scheme, in which inference is performed jointly on a batch of query samples. Transduction leverages the statistics of the unlabeled query samples, yielding notable performance gains, and have triggered a large body of recent works in few-shot learning, with various methods and mechanisms \cite{LIULB2019, TIM_NEURIPS2020, ILPC_ICCV2021, PADDLE_NEURIPS2022, PROTOLP_CVPR2023, AMWACV2024, HELACVPR24}, as detailed in Section~\ref{sec:related}. Generally, transductive methods are built upon optimizing objective functions that integrate clustering terms, either discriminative as in information maximization \cite{TIM_NEURIPS2020,ALPHATIM_NEURIPS2021} or generative as in probabilistic K-means \cite{PADDLE_NEURIPS2022,EMDIRICHLET_CVPR2024}. However, many among these assume a perfectly balanced 
class distribution within the query set, and incorporate terms in the objective function or constraints that enforce a class-balance prior. 
The latter could limit the applicability of these methods, whose performances were shown to drop significantly when dealing with imbalanced query sets \cite{TAFFSL2020, ALPHATIM_NEURIPS2021}.

Several recent attempts tackled this limitation, to handle more realistic scenarios \cite{ALPHATIM_NEURIPS2021, PADDLE_NEURIPS2022, TAO2023, AMWACV2024, EMDIRICHLET_CVPR2024}. Yet, these methods introduce key hyper-parameters, which control the prediction statistics of the unlabeled query set, such as the level of class balance. Such hyper-parameters are empirically grid-searched over pre-defined sets of values using validation data, and their optimal configurations may vary substantially with the target dataset and pre-training model~\cite{pt-map_ICANN2021}. To illustrate this, we depict in Fig.  \ref{fig:acc_lambda} the accuracy as a function of a class-balance hyper-parameter for the recent method in \cite{EMDIRICHLET_CVPR2024}. 
One may observe that this hyper-parameter has a crucial effect on the performance, and its optimal value might vary by orders of magnitude across the target datasets. The issue is further compounded by additional hyper-parameters, which makes
intensive empirical grid searches for the hyper-parameters over validation data and pre-defined intervals of values both sub-optimal and computationally intractable. Therefore, we advocate and introduce the unrolling paradigm (also called ``learning to optimize") in the context of few-shot learning, which enables to learn efficiently and effectively a set of optimized hyper-parameters.
Specifically, our main contributions could be summarized as follows:
\begin{enumerate}
\item We study a generalization of the ubiquitous Expectation-Maximization (EM) algorithm, in which we make two hyper-parameters controlling prediction statistics -- class balance and prediction entropy -- explicit\footnote{Those hyper-parameters are implicit (hidden) in the standard EM formulation.}. Our generalization encompasses several existing transductive few-shot methods as particular cases and, more importantly, enables to learn these crucial hyper-parameters through the proposed unrolling strategy.   
\item We unroll the generalized EM optimizer into a neural network architecture, mapping each of its iterations to a layer and learning the introduced hyper-parameters over validation data. Our unrolling approach offers greater flexibility in optimizing the hyper-parameters, allowing them to vary across the network's layers.
To the best of our knowledge, this study is the first to investigate unrolling in transductive few-shot learning.
\item We design our unrolling architecture in way that covers various statistical assumptions and pre-training paradigms: (i) Gaussian for 
vision-only classifiers and (ii) Dirichlet for vision-language models.
\item We report comprehensive experiments, which cover a breadth of fine-grained downstream image classification tasks and different pre-training models, showing substantial gains over recent state-of-the-art methods.  
\end{enumerate}

%% file: sec/2_related_work.tex
\section{Related works}
\label{sec:related}
\textbf{Few-shot classification with vision-only models:}
Few-shot classification using vision models has been widely explored in the literature, leading to the development of various methods. The first category, inductive methods, predicts the class of each test sample (in the query set) independently from others \cite{SHELL2022, LUO2023}. In contrast, the second category, transductive methods, which has gained increasing attention in recent years, involves jointly predicting classes for a batch of test samples in each few-shot task. Numerous research efforts in transductive methods have leveraged concepts like clustering \cite{PADDLE_NEURIPS2022, AMWACV2024}, label propagation \cite{LIULB2019, PROTOLP_CVPR2023}, information maximization \cite{TIM_NEURIPS2020, ALPHATIM_NEURIPS2021}, optimal transport \cite{ILPC_ICCV2021, CTICCV2023}, prototype estimation \cite{PROTOLP_CVPR2023, BD-CSPN_ECCV2020}, and variational networks \cite{VAR2023TMLR, HELACVPR24}, among others. Studies have shown that transductive methods significantly outperform inductive approaches, achieving accuracy gains of up to 15\%
as reported in several evaluations \cite{TIM_NEURIPS2020, PADDLE_NEURIPS2022}.

\noindent \textbf{Few-shot classification with vision-language models:}
Contrastive Vision-Language Pre-training (CLIP) has recently emerged as an effective model for enhancing various vision tasks through visual-text pairs. By learning transferable visual models with natural language supervision, CLIP demonstrates strong performance in zero-shot classification by matching the image features to text embeddings of novel classes \cite{RADFORD2021}. To further enhance its classification capabilities, several studies have extended CLIP to the few-shot setting \cite{STRONGERLP24, CoOp22, PLOT23, KgCoOp23, PROGRAD23}.
While the linear probe  model~\cite{RADFORD2021} 
trains a logistic regression classifier using CLIP image features, the authors of \cite{STRONGERLP24} proposed a 
generalization of this baseline, modeling the classifier weights as learnable functions of the visual 
prototypes and text embeddings. Context Optimization (CoOp) \cite{CoOp22} and its extended versions \cite{KgCoOp23, PLOT23, PROGRAD23} 
have been developed based on the concept of prompt learning. While CoOp \cite{CoOp22} aims to model context in prompts using 
continuous representations, a knowledge-guided CoOp is proposed in \cite{KgCoOp23} to enhance the 
generalization ability by minimizing the discrepancy between the learnable prompts and the original ones. 
In \cite{PLOT23}, the authors learn multiple prompts, describing the characteristics of each class, through 
the minimization of an optimal-transport distance. Unlike prompt learning methods, which fine-tune the input text, 
another family of methods, referred to as \textit{adapters}, aims to transform the visual or text encoders, such as \cite{TIPADAP22, CLIPADAP23}. For instance, TIP-Adapter \cite{TIPADAP22} adds a non-parametric adapter to the weight-frozen CLIP model, and updates the prior knowledge encoded in CLIP by feature retrieval. CLIP-Adapter \cite{CLIPADAP23} introduces a multi-layered perceptron to learn new features, and combine them with the original CLIP-encoded features via residual connections. \\
It is worth noting that all the aforementioned methods belong to the inductive family. However, unlike vision-only models, 
the transductive few-shot setting is still not well investigated in the context of CLIP. To the best of our knowledge, the only transductive 
few-shot CLIP method is the one very recently proposed in~\cite{EMDIRICHLET_CVPR2024}, which, inspired by the Expectation Maximization algorithm, 
relies on the Dirichlet distribution to model the data. Moreover, as reported in \cite{EMDIRICHLET_CVPR2024}, and unlike the behaviors observed with vision-only models, recent transductive few-shot methods do not always outperform their inductive counterparts with CLIP. This suggests that there is a need to further investigate transductive methods in the context of vision-language models, due to their aforementioned advantages with respect to inductive methods.

\noindent \textbf{Class-balance and hyperparameters setting:}
\label{sec:class_balance}
Hyperparameter optimization has a large impact on the performance of a given iterative algorithm or learning model, and is generally tackled through greedy or gradient-based methods \cite{Grad_Hyperparams_NEURIPS2021}. In the context of transductive few-shot learning, most of the existing methods have been designed for perfectly balanced query 
sets (i.e., uniform class distribution), and have shown drops in performances under class-imbalanced 
settings. To address this flaw, various methods have been developed in the last years \cite{ALPHATIM_NEURIPS2021, 
PADDLE_NEURIPS2022, TAO2023, AMWACV2024, EMDIRICHLET_CVPR2024}. 
For instance, in \cite{TAO2023} the categorical probability of each query sample is regularized to 
quantify the difference between the class marginal distribution and the uniform one. The works in 
\cite{ALPHATIM_NEURIPS2021, PADDLE_NEURIPS2022, EMDIRICHLET_CVPR2024, AMWACV2024} explored various weighted terms, which are added 
to the objective functions, to mitigate the effect of the class-balance bias. To this end, 
some hyper-parameters are introduced, to weigh to contributions of such terms in the objective function. Unlike \cite{PADDLE_NEURIPS2022},  
where the introduced hyper-parameter has been theoretically chosen in order to compensate for the hidden class-balance bias in the used clustering objective, the hyper-parameters are often set in an empirical manner based on the validation classes of each dataset. Thus, a set of hyperparameter values (in a given range) are evaluated, and the ones yielding the highest accuracy are selected.

%% file: sec/3_methodology.tex
\section{Proposed methodology}
\label{sec:method}
\subsection{Generalized EM algorithm}

\textbf{Preliminaries}: Let us first introduce the notations to formulate our transductive few-shot inference. Let \( \{\bs{z}_n\}_{1 \leq n \leq N} \) denote the set of feature vectors extracted from a pre-training network, and which are to be classified, with $N$ the total number of samples within a given task. 
The whole dataset contains \( K \) distinct classes, whereas the number of randomly sampled classes present in each mini-batch task might be much smaller than $K$. This subset of sampled classes may also vary across mini-batches, and the method does not assume any prior knowledge on the specific classes that may appear in each mini-batch.

Moreover, for a given few-shot task, let us denote by $ \support \subset \{1, \dots, N\}$ and $\query = \{1, \dots, N\} \setminus  \support$ the indices of samples within the support (labeled) and query (unlabeled) mini-batch sets, respectively. For every $n \in \support$, $\bs{y}_n = (y_{n,k})_{1 \leq k \leq K} \in \{0,1\}^K$ are the one-hot-encoded labels, such that, for every $k \in \{1, \dots, K\}$, $y_{n,k}=1$ if the $n$-th sample belongs to class $k$, and $y_{n,k}=0$ otherwise. 

Finally, 
given the extracted feature vectors $\bs{z}_n$, we assume that the data probability distribution knowing its class $k$ is modeled by a given law, whose probability density function (pdf) is denoted by $\mathrm{p} \left( \boldsymbol{z}_n ~|~ \boldsymbol{\theta}_k \right )$ and characterized by a vector of parameters $\bs{\theta}_{k}$. This means that the global distribution of $\boldsymbol{z}_n$
is a mixture of these pdfs.

\noindent \textbf{Problem formulation}: Our goal is to identify the classes of the unlabeled samples in the query set by optimizing a general clustering objective function, while embedding supervision constraints from the few labeled samples within the support set. We do so by unrolling iterative block-coordinate optimizers of the objective functions over two sets of variables: 
\begin{itemize}
\item Soft assignment vectors $\bs{u} = (\bs{u}_n)_{1 \leq n \leq N} \in (\Delta_K)^N$, where $\Delta_K$ is the probability simplex of $\mathbb{R}^{K}$. For every 
$n\in \{1,\ldots,N\}$, $\bs{u}_n = (u_{n,k})_{1 \leq k \leq K}$ where, for every $k\in \{1,\ldots,K\}$, $u_{n,k}$ can be interpreted as the probability that the 
$n$-th sample belongs  to class $k$. These probabilities have to be determined for the query set samples.
\item Feature distribution parameters $\bs{\theta}=(\bs{\theta}_k)_{1 \leq k \leq K}$. 
\end{itemize}

Consider the following general probabilistic clustering problem:
\begin{alignat}{2}\label{eq:opt_prob}
    &\underset{\bs{u}, \bs{\theta}}{\mathrm{minimize}}\,   &&  \mathcal{L}(\bs{u}, \bs{\theta}) + \lambda  \Psi(\bs{u}) + T \Phi(\bs{u}),\\
    &\text{subject to} && \, \bs{u}_n \in \Delta_K \quad \forall n \in \query, \nonumber\\
    & && \, u_{n,k} = y_{n, k} \quad \forall n \in \support, \, \forall k \in \{1, \dots, K\}. \nonumber
\end{alignat}
where weighing factor $\lambda$ and temperature scaling $T$ are learnable optimized hyper-parameters, which we will estimate through the proposed unrolling strategy (Section~\ref{sec:unrolling_EM}), and terms $\mathcal{L}$, $\Psi$ and $\Phi$ are detailed in the following. 
\begin{itemize}
\item The first term in objective function \eqref{eq:opt_prob} 
is the negative log-likelihood of the feature vectors:
\begin{align}
\label{log-likelihood}
\mathcal{L}(\bs{u},\bs{\theta})= - \sum_{n=1}^{N}\sum_{k=1}^{K}u_{n,k}\mathrm{ln}(\p(\bs{z}_n|\bs{\theta}_k))
\end{align}
This general log-likelihood model fitting term is well known in the context of clustering methods \cite{Kearns-UAI-97,Boykov-ICCV-05}. In fact, it generalizes the standard $K$-means clustering objective to arbitrary distributions\footnote{K-means corresponds to choosing the multivariate Gaussian distribution, with identity covariance matrix, for parametric density $\p(\bs{z}_n|\bs{\theta}_k)$.}.
It is well-known that minimizing \eqref{log-likelihood} has an inherent bias towards class-balanced clustering \cite{Kearns-UAI-97,Boykov-ICCV-05,PADDLE_NEURIPS2022,EMDIRICHLET_CVPR2024}.

\item The second term in \eqref{eq:opt_prob} controls the partition complexity of the model, penalizing the number of non-empty clusters in the solution.
It corresponds to the Shannon entropy of class distribution $(\pi_k)_{1 \leq k \leq K}$, 
defined as
\begin{equation}
\label{eq:regularization_mdl}
  \Psi(\bs{u})=  - \sum_{k=1}^K \pi_k \ln \pi_k,
\end{equation}
where $\pi_k= \frac{1}{|\query|} \sum_{n\in \query} u_{n, k}$ is the proportion of query samples within class $k$, i.e., the empirical estimate of the marginal 
probability of class $k$. 

\noindent\textbf{Class-balance hyper-parameter ($\lambda$)}: The marginal entropy in \eqref{eq:regularization_mdl} mitigates the class-balance bias of the log-likelihood clustering term in \eqref{log-likelihood}.  It reaches its minimum for the extremely imbalanced solution in which all data samples are assigned to a single cluster, and its maximum for a perfectly balanced clustering. Hence, clearly, hyper-parameter $\lambda$ controls the level of class balance in the solution. As depicted in Fig. \ref{fig:acc_lambda}, this hyper-parameter has a crucial effect on the performance, and its optimal value might vary by orders of magnitude from one dataset to another. This makes exhaustive grid searches for optimal $\lambda$ over validation sets intractable computationally, which motivates learning this hyper-parameter through our unrolling strategy, as described in Section \ref{sec:unrolling_EM}. 

\item The third term in \eqref{eq:opt_prob} is an entropic barrier, enabling to soften assignments \( u_{n,k} \), while imposing a non-negativity constraint on each of them. It is given by
\begin{align}
\label{entropic-barrier}
\Phi(\bs{u}) = \sum_{n=1}^N \sum_{k=1}^K u_{n,k} \ln u_{n,k}.
\end{align}
\noindent\textbf{Temperature scaling hyper-parameter ($T$)}: Weighting factor $T$ in \eqref{eq:opt_prob} controls the trade-off 
between the clustering term and the entropic barrier in \eqref{entropic-barrier}, i.e., the level of softness of assignments \( u_{n,k} \). Therefore, this hyper-parameter has an important effect on performances and on the marginal class probabilities appearing in the class-balance term in \eqref{eq:regularization_mdl}, which motivates learning it.   
As described in Section \ref{sec:unrolling_EM}, our unrolling algorithm enables to cope with this extra parameter.
\end{itemize}

\noindent \textbf{Link to EM and other transductive few-shot methods}: 
We examine solving problem \eqref{eq:opt_prob} with an iterative block-coordinate descent algorithm (see Algorithm \ref{algo:Gen_EM}), which alternates two steps, one updating distribution parameters $\bs{\theta}_k^{(\ell)}$ and the other optimizing over class assignments $\bs{u}_n^{(\ell)}$, at each iteration $\ell$. For the $\bs{u}$-step, and due to the nonconvexity of $\Psi$, we proceed with a Majorization-Minimization (MM) strategy to minimize a surrogate convex function
with respect to $\bs{u}$ at each iteration. 
In addition, if $\p(\cdot|\bs{\theta}_k)$ is assumed to 
belong to the exponential family and $\bs{\theta}_k$ are its canonical parameters,
the estimation w.r.t. $\bs{\theta}$, with $\bs{u}$ fixed, is a convex problem. We provide further details on this general iterative optimization scheme in Appendix \ref{app:det_min_steps}.
It is worth noting that, when $T=1$ and the data is modeled by the Dirichlet distribution, we recover the recent transductive few-shot learning algorithm in \cite{EMDIRICHLET_CVPR2024}. Interestingly, when $T=1$
and $\lambda = |\query|$, we recover the well-known Expectation-Maximization (EM) algorithm for
estimating the parameters of mixture of distributions; see Proposition~1 in \cite{EMDIRICHLET_CVPR2024}.
Therefore, Algorithm \ref{algo:Gen_EM} could be viewed as a generalized EM (GEM), in which hyper-parameters $\lambda$ and $T$ control the class balance as well as prediction softness, and could be made learnable. 
\begin{algorithm*}
\caption{GEM based few-shot classification algorithm \label{algo:Gen_EM}}
\begin{algorithmic}
\STATE \textbf{Input:} Compute $\bs{z}_n$ for the dataset samples, initialize $\boldsymbol{u}_{n}^{(0)}$  and $\boldsymbol{\theta}_{k}^{(0)}$, and fix the number of iterations $L$,
\FOR{$\ell = 0,1,\ldots,L-1$} 
\STATE \textcolor{gray}{// Update Distribution parameters for each class using a given estimation algorithm (denoted here by ``DP\_est'')}
\STATE $\displaystyle\bs{\theta}_k^{(\ell +1)} = \mathrm{DP\_est}(\bs{u}_{\cdot, k}^{(\ell)}, \bs{\theta}_k^{(\ell)}) $,  \quad $\forall k \in\{1, \dots, K\},$\\
\STATE \textcolor{gray}{// Update class proportions}
\STATE $\displaystyle\pi_{k}^{(\ell+1)} = \frac{1}{|\query|} \sum_{n\in \query} u_{n, k}^{(\ell)}$,  \quad $\forall k \in\{1, \dots, K\},$\\
\STATE \textcolor{gray}{// Update assignment vectors for all query samples}
\STATE $\displaystyle \bs{u}_n^{(\ell +1)} =  \text{\normalfont softmax}\left( \frac{1}{T} \left(\ln \mathrm{p} \left( \boldsymbol{z}_{n} ~|~ {\bs{\theta}_k^{(\ell+1)}}  \right)+ \frac{\lambda}{|\query|}\ln(\pi_{k}^{(\ell+1)}) \right)_{ k}  \right) $, \quad $\forall n \in \query.$ 
\ENDFOR
\end{algorithmic}
\end{algorithm*}
\begin{figure*}[!h]
\centering
\includegraphics[width=0.70\textwidth]{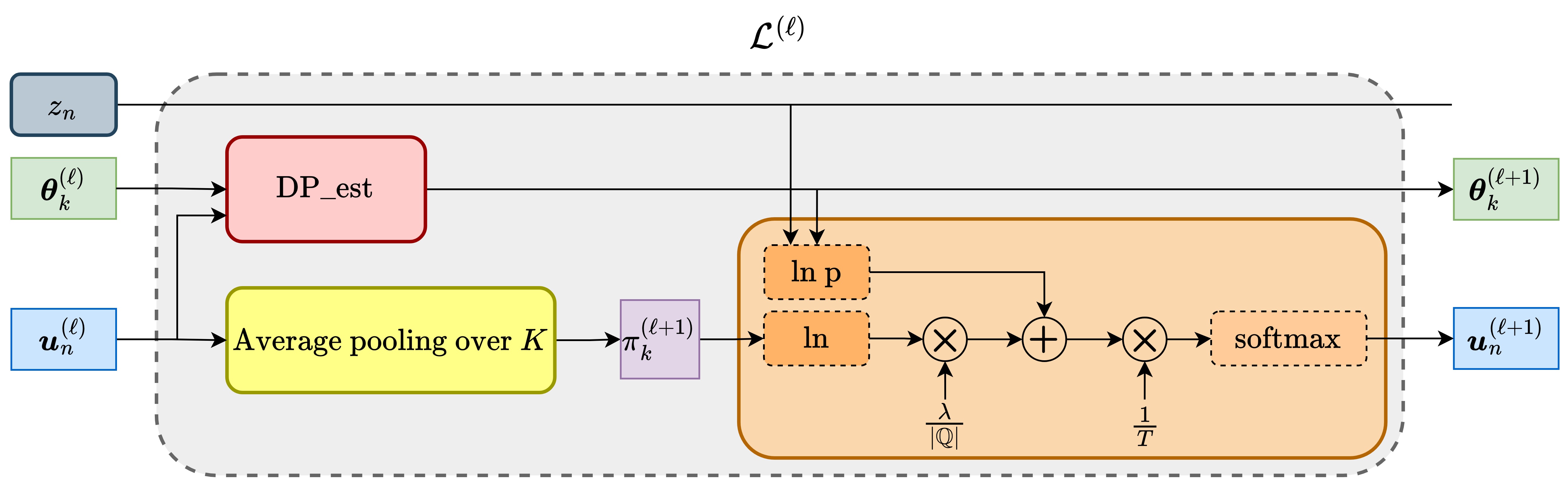} \vspace{-0.2cm}
\caption{An overview of the unrolled GEM algorithm for a given iteration. Each iteration $\ell$ corresponds to a network layer $\mathcal{L}^{(\ell)}$.
Each layer depends on the vector of hyperparameters $(\lambda^{(\ell)}, T^{(\ell)})$.}
\label{fig:UNEM_lay}
\end{figure*}

\noindent\textbf{Examples of data models}: While a broad range of distribution models in the exponential family could be adopted, we focus in this paper on two popular ones. 
\begin{itemize}
\item The first one is the Gaussian distribution, which is commonly used in standard clustering and transductive few-shot-methods applied to vision models \cite{MPP2021, ALPHATIM_NEURIPS2021,PADDLE_NEURIPS2022}. By assuming a Gaussian distribution with mean $\bs{\theta}_k$ and identity covariance matrix, the pdf $\mathrm{p} (\boldsymbol{z}_{n} | \bs{\theta}_k )$ reads as follows
\begin{equation}
\label{eq:gauss_distribution}
\mathrm{p} (\boldsymbol{z}_{n} | \bs{\theta}_k ) \propto  \exp \left(-\frac{1}{2}\|\bs{z}_n-\bs{\theta}_k \|^2 \right).
\end{equation}
\item The second one is the Dirichlet distribution, which has recently shown good modelling performance in the context of transductive few-shot for vision-language models such as CLIP \cite{EMDIRICHLET_CVPR2024}. For $\bs{z}_n=(z_{n,i})_{1 \leq i \leq K}$ and  $\bs{\theta}_k=(\theta_{k,i})_{1 \leq i \leq K}$, the associated pdf is given by
\begin{equation}
\label{eq:dirichlet_distribution}
\mathrm{p} \left( \boldsymbol{z}_n ~|~ \boldsymbol{\theta}_k \right ) = \frac{1}{\mathcal{B}(\boldsymbol{\theta}_k)} \prod_{i=1}^K z_{n,i}^{\theta_{k, i} -1} \, \mathbbm{1}_{\boldsymbol{z}_n \in \Delta_K},
\end{equation}
where the normalization factor $\mathcal{B}(\boldsymbol{\theta}_k)$ is
\begin{equation}
   \mathcal{B}(\boldsymbol{\theta}_k) = \frac{\prod_{i=1}^K \Gamma(\theta_{k,i}) }{\Gamma\left(\sum_{i=1}^K \theta_{k,i}\right)}, 
\end{equation}
and $\Gamma$ denotes the Gamma function.
\end{itemize}
The necessary details related to feature representations in the cases of vision-only and vision-language models, distribution parameter estimation, and the resulting GEM algorithms for Gaussian and Dirichlet laws are provided in Appendices \ref{app:EM_Gaus} and \ref{app:EM_Dir}, respectively.
\subsection{UNrolled EM architecture (UNEM)} 
\label{sec:unrolling_EM}
\begin{figure*}[!h]
  \centering
 \includegraphics[width=0.7\textwidth]{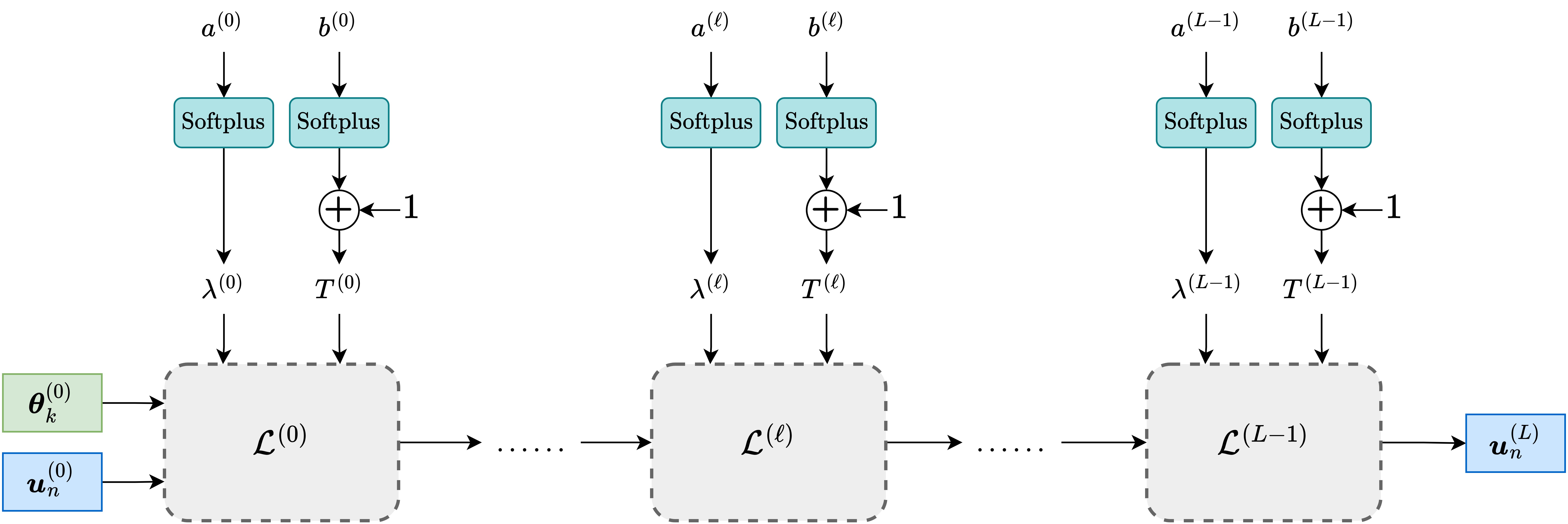} \vspace{-0.2cm}
    \caption{Overall architecture of the designed UNEM.}
    \label{fig:UNEM_over_arch}
\end{figure*}
\textbf{Overview}: As discussed in Section \ref{sec:class_balance}, the choice of the optimal hyper-parameter, which controls the level of class balance, is a 
difficult task, often performed manually in the existing transductive few-shot classification methods. 
In addition to the 
class-balance parameter $\lambda$, Algorithm \ref{algo:Gen_EM} involves an additional temperature parameter $T$, 
which makes hyper-parameter setting even more challenging. For this reason, we propose in this paper to resort to the unrolling (called also \textit{learning to optimize}) paradigm \cite{Eldar_2021}, which enables to learn efficiently a set of optimized hyper-parameters. It is important to note that unrolling iterative optimization algorithms has found successful applications in diverse 
signal and image processing tasks \cite{MCC_2020, Eldar_2021, CHENTMLR2022}. However, to the best of our knowledge, this work is the first to leverage the unrolling paradigm for hyper-parameter optimization in a few-shot learning context. 
The main idea behind the learning-to-optimize paradigm 
is to map each iteration of a given optimization algorithm to a network layer, stack all layers together, and 
view the hyper-parameters to be optimized as the network's learnable parameters.
More precisely, to unroll our generalized EM algorithm, the number of iterations $L$ is used as the number of layers 
for the neural network architecture. Thus, each iteration $\ell \in \{0,1,\ldots,L-1\}$ is associated with a tailored layer 
$\mathcal{L}^{(\ell)}$, which performs the update rules defined in Algorithm~\ref{algo:Gen_EM}:
\begin{equation}
  \left( \bs{\theta}_k^{(\ell+1)}, \displaystyle\pi_k^{(\ell+1)}, \bs{u}_n^{(\ell+1)} \right)=\mathcal{L}^{(\ell)}\left( \bs{\theta}_k^{(\ell)}, \displaystyle\pi_k^{(\ell)}, 
\bs{u}_n^{(\ell)}; \lambda^{(\ell)}, T^{(\ell)} \right)
\end{equation}
where $(\lambda^{(\ell)},T^{(\ell)})_{0\leq \ell \leq L-1}$ is the vector of hyper-parameters to be learned. 
This leads to the UNrolled EM (UNEM) model shown in Fig. \ref{fig:UNEM_lay}, which depicts the inputs and outputs of a given layer $\mathcal{L}^{(l)}$ as well as its three main blocks for the update rules. 

\noindent \textbf{Learned hyper-parameters}: Instead of a restricted number of handcrafted hyper-parameters 
(as often used in iterative algorithms), our unrolled algorithm offers more flexibility, enabling adaptation of the hyper-parameters along the processing workflow.
This means that
$L$ vectors of hyper-parameters $(\lambda^{(\ell)},T^{(\ell)})_{0\leq \ell \leq L-1}$ could be learned and
applied at each layer of the network. During the training of the unrolled model, we fulfill some design constraints, as described in more details in the following.\\
More specifically, to ensure the non-negativity of hyper-parameter $\lambda^{(\ell)}$, we propose to express it as a Softplus function, which could be seen as a smooth approximation of the RELU activation function, yielding
\begin{equation}
\lambda^{(\ell)}=\operatorname{Softplus}(a^{(\ell)})=\log(1+\exp(a^{(\ell)}))
\end{equation}
where $a^{(\ell)}$ represents a learnable parameter for the unrolled architecture.

Regarding temperature scaling $T^{(\ell)}$, we observed in our experiments that imposing the non-negativity constraint alone might be a source of instability, resulting in values very close to zero and vanishing gradient issues during the training 
of the unrolled architecture. To circumvent this problem, we impose a lower bound equal to 1 on the temperature scaling:
\begin{equation}
T^{(\ell)}=1+\operatorname{Softplus}(b^{(\ell)})=1+\log(1+\exp(b^{(\ell)}))
\end{equation}
where $b^{(\ell)}$ denotes a parameter that needs to be learned during the training of the unrolled model. 

\noindent \textbf{Overall architecture and training approach}: Based on the previous considerations, the overall architecture of UNEM can be summarized as the composition of the $L$ layers $\mathcal{L}^{(L-1)} \circ \ldots \circ \mathcal{L}^{(0)}$. This architecture, depicted in Figure~\ref{fig:UNEM_over_arch}, illustrates: (i) the required inputs $\bs{\theta}_k^{(0)}$ and $\bs{u}_n^{(0)}$ as well as the feature vectors $\bs{z}_n$ (which are fed in all 
layers), (ii) the cascaded layers with their associated vector of parameters $\bs{w}=(\lambda^{(\ell)},T^{(\ell)})_{0 \leq \ell \leq L-1}$ satisfying the aforementioned constraints, and (iii) the key output $\bs{u}_{n}^{(L)}$ representing the class assignment vectors. 

The resulting neural network architecture is trained by minimizing a standard cross-entropy loss on a validation set:
\begin{equation}
 \mathrm{L}_\mathrm{c}(\bs{w})=\sum_{n \in \query}\sum_{k=1}^{K} y_{n,k}\log(u^{(L)}_{n,k}).
\end{equation}



%% file: sec/4_experiments.tex
\section{Experiments} 
\label{sec:experiments}
The effectiveness of the proposed approach is validated in both vision-only and vision-language transductive few-shot learning settings. We will designate our proposed 
unrolled  methods by UNEM-Gaussian and UNEM-Dirichlet, respectively.
After describing the experimental settings, this section is structured into two main sections discussing our experiments in both evaluation scenarios. 
\begin{table*}[!h]
    \centering
    \scalebox{0.82}
    {
    \begin{tabular}{l c ccc ccc}
        \toprule
        \multirow{2}{*}{\textbf{Method}} & \multirow{2}{*}{\textbf{Backbone}} & \multicolumn{3}{c}{\textbf{\textit{mini}-ImageNet} (\textcolor{blue}{$K=20$})} & \multicolumn{3}{c}{\textbf{\textit{tiered}-ImageNet (\textcolor{blue}{$K=160$})}} \\
        \cmidrule(lr){3-5} \cmidrule(lr){6-8}
         & & 5-shot & 10-shot & 20-shot & 5-shot & 10-shot & 20-shot \\
        \bottomrule
        Baseline \cite{BASELINE_ICLR2019} & \multirow{9}{*}{ResNet-18} & 55.4 & 62.1 & 67.9 & 29.7 & 36.3 & 42.2 \\
        LR+ICI \cite{LR+ICI_CVPR2020} & & 55.4 & 62.1 & 68.1 & – & – & – \\
        BD-CSPN \cite{BD-CSPN_ECCV2020} & & 49.8 & 54.6 & 56.5 & 11.4 & 11.0 & 11.7 \\
        PT-MAP \cite{pt-map_ICANN2021} & & 25.7 & 27.2 & 28.4 & 5.2 & 6.0 & 6.6 \\
        LaplacianShot \cite{LAPLACIANSHOT_ICML2020} & & 57.9 & 64.2 & 68.3 & 29.6 & 35.4 & 39.1 \\
        TIM \cite{TIM_NEURIPS2020} & & \textbf{66.8} & 69.9 & 70.8 & 29.3 & 28.7 & 27.8 \\
        $\alpha$-TIM \cite{ALPHATIM_NEURIPS2021} & & 66.7 & 71.0 & 73.9 & 43.8 & 48.3 & 51.9 \\
        $\alpha$-AM \cite{AMWACV2024} & & 64.4 & 67.8 & 70.1 & - & - & - \\
        PADDLE \cite{PADDLE_NEURIPS2022} & & 62.9 & 73.5 & 79.8 & 45.4 & 61.4 & 70.6 \\
        \rowcolor[HTML]{ffa6a6} 
        \textbf{UNEM-Gaussian} & & 66.4 & \textbf{75.6} & \textbf{80.4} & \textbf{52.3} & \textbf{65.7} & \textbf{73.2}\\
        \midrule
        
        Baseline \cite{BASELINE_ICLR2019} & \multirow{9}{*}{WRN28-10} & 59.0 & 65.7 & 72.1 & 31.9 & 39.0 & 45.6 \\
        LR+ICI \cite{LR+ICI_CVPR2020} & & 58.8 & 65.7 & 72.0 & – & – & – \\
        BD-CSPN \cite{BD-CSPN_ECCV2020} & & 51.1 & 55.5 & 58.4 & 18.0 & 18.4 & 18.2 \\
        PT-MAP \cite{pt-map_ICANN2021} & & 26.5 & 28.0 & 29.3 & 5.2 & 6.0 & 6.6 \\
        LaplacianShot \cite{LAPLACIANSHOT_ICML2020} & & 61.0 & 66.8 & 71.0 & 31.4 & 37.3 & 41.5 \\
        TIM \cite{TIM_NEURIPS2020} & & \textbf{72.1} & 74.9 & 76.2 & 36.1 & 39.0 & 38.5 \\
        $\alpha$-TIM \cite{ALPHATIM_NEURIPS2021} & & 71.5 & 75.2 & 78.3 & 45.8 & 51.4 & 55.2 \\
        $\alpha$-AM \cite{AMWACV2024} & & 68.2 & 71.3 & 73.3 & - & - & - \\
        PADDLE \cite{PADDLE_NEURIPS2022} & & 62.6 & 73.0 & 79.2 & 43.9 & 59.4 & 69.9 \\

        \rowcolor[HTML]{ffa6a6}
        \textbf{UNEM-Gaussian} & & 71.6 & \textbf{79.2} & \textbf{83.7} & \textbf{54.1} & \textbf{66.8} & \textbf{74.7} \\ 
        \bottomrule
    \end{tabular} 
    }
    \caption{Comparison of the proposed UNEM-Gaussian with respect to state-of-the-art methods on \textit{mini}-Imagenet and \textit{tiered}-Imagenet. The metric is accuracy (in percentage). Results are averaged across 1,000 tasks. Results marked with ’-’ were intractable to obtain.}
    \label{tab:vis_main_exp}
\end{table*}

\begin{table}[!h]
    \centering
      \scalebox{0.82}
      {
    \begin{tabular}{l ccc}
        \toprule
        \multirow{2}{*}{\textbf{Method}} & \multicolumn{3}{c}{\textbf{CUB} (\textcolor{blue}{$K=50$})} \\
        \cmidrule(lr){2-4} 
          & 5-shot & 10-shot & 20-shot \\
        \bottomrule
        Baseline \cite{BASELINE_ICLR2019} & 58.6 & 68.8 & 78.2 \\
        LR+ICI \cite{LR+ICI_CVPR2020} & 49.9	& 55.6 & 58.0 \\
        PT-MAP \cite{pt-map_ICANN2021} & 12.8	& 14.0 & 14.9 \\
        LaplacianShot \cite{LAPLACIANSHOT_ICML2020} & 58.8	& 66.5 & 71.0 \\
        TIM \cite{TIM_NEURIPS2020} & 68.1	& 68.9 & 69.3 \\
        $\alpha$-TIM \cite{ALPHATIM_NEURIPS2021} & 74.3	& 79.3 & 83.6 \\
        $\alpha$-AM \cite{AMWACV2024} & 66.2	& 68.9 & 69.8 \\
        PADDLE \cite{PADDLE_NEURIPS2022} & 71.2 & 81.8 & 86.8 \\
        \rowcolor[HTML]{ffa6a6}
        \textbf{UNEM-Gaussian} & \textbf{78.5} & \textbf{85.3} & \textbf{88.6} \\
        \bottomrule
        \vspace{-0.5cm}
    \end{tabular} 
    }
    \caption{Comparison of the proposed UNEM-Gaussian with respect to state-of-the-art methods on CUB. The metric is accuracy (in percentage). Results are averaged across 1,000 tasks.}
    \label{tab:vis_cub_exp}
\end{table}
\subsection{Experimental settings} \label{sec:exp_settings}
\noindent \textbf{Task generation}: 
We adopt a realistic transductive few-shot evaluation protocol,  which is in line with state-of-the-art protocols  \cite{EMDIRICHLET_CVPR2024,PADDLE_NEURIPS2022,PROTONET,ALPHATIM_NEURIPS2021,LIULB2019}. While $K$ designates the total number of classes in the labeled support set $\support$, $K_{\rm eff}$ (with $K_{\rm eff} \ll K$) denotes the number of effective classes present in the unlabeled query set $\query$. The classes in the query set remain undisclosed during inference, while randomly selecting $|\mathbb{Q}|$ samples. On the other hand, the support set is built by uniformly selecting $s$ images from each of the $K$ classes. In this paper, the few-shot tasks are performed with (i) 5, 10, and 20 shots for vision-only models, and (ii) 4 shots for vision-language models. Moreover, in both of our evaluation scenarios, we used $K_{\rm eff} = 5$ and $|\mathbb{Q}|=75$.  \\
\noindent\textbf{Training specifications}: For fair comparison and reproducibility purposes, the CLIP's pre-trained model has been directly used in the vision-language evaluation scenario, while for the vision-only scenario, the standard pre-trained ResNet-18 and WRN28-10 backbones have been fine-tuned. Their training is performed on the base classes set of each vision-only dataset, using cross-entropy loss with label smoothing set to 0.9 for 90 epochs. The learning rate is set to 0.1 and then decayed by a factor of 10. \\
On the other hand, the training of our unrolled architecture, composed of 10 layers (i.e.,  $L=10$), is performed on several tasks sampled from the validation set of each dataset.
This implies that the proposed solution is highly economical in terms of parameters, typically requiring only a few tenths, compared to standard neural network architectures.
Let us recall that the validation set is often used to select the hyperparameters in recent state-of-the-art transductive methods as mentioned in Section \ref{sec:class_balance}. Specifically, for the vision-only (resp. vision-language) models, the training of the unrolled architecture is carried out on 1,000 (resp. 100) tasks, while using an initial learning rate of 0.1 (resp. 0.5) with a decay factor of 0.5, a number of epochs equals to 80, and ADAM as optimizer. The trainable hyper-parameters are initialized to the default values set in \cite{PADDLE_NEURIPS2022} (i.e., $\lambda=|Q|$ for the vision-only models) and \cite{EMDIRICHLET_CVPR2024} (i.e.,  $\lambda=\frac{K}{K_{\rm eff}}$ for the vision-language models, with $T=1$). Our architecture is implemented in Pytorch (version 2.3.0) and run on NVIDIA QUADRO RTX8000 (with 48 GB of memory).

\begin{table*}[!h]
    \centering
    \scalebox{0.75}{
    \begin{tabular}{c l   cccccccccccc}
         & \textbf{Method}
         & \rotatebox{60}{Food101} 
         & \rotatebox{60}{EuroSAT} 
         & \rotatebox{60}{DTD} 
         & \rotatebox{60}{OxfordPets} 
         & \rotatebox{60}{Flowers102} 
         & \rotatebox{60}{Caltech101} 
         & \rotatebox{60}{UCF101} 
         & \rotatebox{60}{FGVC Aircraft} 
         & \rotatebox{60}{Stanford Cars} 
         & \rotatebox{60}{SUN397}
         & \rotatebox{60}{ImageNet} 
         & \rotatebox{60}{Average} \\
        \bottomrule
         
        \multirow{2}{*}{\rotatebox{90}{\textcolor{blue}{Ind.}}} & Tip-Adapter \cite{TIPADAP22}& 76.7 & \textbf{72.5} & 54.7 & 86.4 & 83.2 & 88.8 & 72.1 & 23.7 & 63.9 & 66.7 & 62.7 & 68.3\\

        & CoOp \cite{CoOp22}& 76.3 & 63.2 & 52.2 & 86.2 & 81.0 & 87.7 & 67.0 & 22.2 & 61.3 & 63.4 & 59.9 & 65.5\\
        \bottomrule
        
        \multirow{6}{*}{\rotatebox{90}{\textcolor{blue}{Transd.}}} & BDSCPN \cite{BD-CSPN_ECCV2020} & 74.7 & 46.1 & 45.2 & 81.3 & 74.2 & 82.0 & 59.0 & 18.0 & 48.1 & 54.5 & 49.2 & 57.5\\

        & Laplacian Shot \cite{LAPLACIANSHOT_ICML2020} & 76.6 & 53.0 & 52.6 & 88.4 & 85.5 & 86.8 & 67.0 & 22.2 & 60.4 & 63.8 & 56.3 & 64.8\\

        & $\alpha$-TIM \cite{ALPHATIM_NEURIPS2021} & 66.1 & 46.1 & 45.3 & 87.1 & 79.1 & 83.3 & 59.4 & 20.4 & 53.4 & 53.4 & 42.7 & 57.8\\

        & PADDLE \cite{PADDLE_NEURIPS2022} & 71.8 & 45.9 & 50.0 & 84.7 & 82.3 & 81.9 & 63.7 & 21.3 & 56.1 & 60.6 & 52.1 & 60.9\\

        & EM-Dirichlet \cite{EMDIRICHLET_CVPR2024} & 88.7 & 50.8 & 62.6 & 92.5 & 91.3 & 90.1 & 76.1 & 24.9 & 73.5 & 80.9 & 78.4 & 73.6\\
        \rowcolor[HTML]{ffa6a6}
        & \textbf{UNEM-Dirichlet} & \textbf{91.4} & 53.8 & \textbf{65.3} & \textbf{96.0} & \textbf{95.6} & \textbf{93.4} & \textbf{78.5} & \textbf{30.4} & \textbf{80.0} & \textbf{88.5} & \textbf{83.1} & \textbf{77.8}\\ \vspace{-0.5cm}

    \end{tabular}
    }
    \caption{\normalsize{Comparison of the proposed UNEM-Dirichlet with respect to the state-of-the-art methods on 11 different datasets. The metric is accuracy (in percentage). Results are averaged across 1,000 tasks.}}
    \label{tab:vlm_main_exp}
\end{table*}
\subsection{UNEM-Gaussian in vision-only few-shot setting}
Let us recall that the Gaussian distribution is commonly used in standard clustering and transductive few-shot-methods applied to vision-only models \cite{MPP2021, ALPHATIM_NEURIPS2021,PADDLE_NEURIPS2022}. For this reason, we will focus here on the proposed UNEM-Gaussian approach. 

\noindent \textbf{Datasets}: The first UNEM-Gaussian architecture is evaluated on the following standard few-shot benchmark datasets: \textit{mini}-ImageNet \cite{mini-imagenet}, \textit{tiered}-ImageNet \cite{tiered-imagenet}, CUB \cite{CUB}. Mini-imagenet has 100 classes split into 64 base classes, 16 validation classes and 20 test classes. The tiered-imagenet has 608 classes instead, from which we follow a standard split with 351 for base training, 97 for validation and 160 for testing. For CUB, we followed the split proposed by \cite{BASELINE_ICLR2019}
which consists of 100 base classes, 50 validation classes and 50 test classes. For each dataset, the feature vectors $\bs{z}_n$ are extracted using the fine-tuned backbones, while applying a scaling parameter $T_z$ to the model's output as described in Appendix~\ref{app:EM_Gaus}. This parameter has also been learned through the unrolling approach. 

\noindent\textbf{Results}: The UNEM-Gaussian architecture has been compared to its original version PADDLE \cite{PADDLE_NEURIPS2022} as well as several state-of-the-art methods. Tables \ref{tab:vis_main_exp} and \ref{tab:vis_cub_exp} presents the results, averaged over 1,000 tasks with 5, 10, and 20 shots. Several observations can be made. First, the proposed UNEM-Gaussian outperforms the original PADDLE algorithm \cite{PADDLE_NEURIPS2022} and the other state-of-the-art methods.
The achieved gains are much higher with fewer number of shots and reach 3.5\% with \textit{mini}-ImageNet, 6.9\% with \textit{tiered}-ImageNet, and 7.3\% with CUB, while using ResNet-18 as backbone. Moreover, when WRN28-10 is used as backbone, it can be noticed that the performance of all methods has been improved, except for PADDLE, which shows results similar to those obtained with ResNet-18 pre-training. This can be explained by the fact that the class-balance parameter $\lambda$ was set in \cite{PADDLE_NEURIPS2022} to $|\query|$ (i.e. 75), which becomes sub-optimal with WRN28-10 features. 
Most importantly, by learning efficiently the hyperparameters, our UNEM-Gaussian achieves higher gain with respect to its original version, yielding an accuracy gain reaching up to 10\% in 5-shot scenario, when WRN28-10 is used as backbone.

\subsection{UNEM-Dirichlet for few-shot CLIP}
Dirichlet distribution has recently demonstrated good modelling performance in the context of transductive few-shot for vision-language models such as CLIP \cite{EMDIRICHLET_CVPR2024}. Thus, we will focus in this second round of experiments on UNEM-Dirichlet approach.

\noindent \textbf{Datasets}: The second unrolled architecture, designated by UNEM-Dirichlet, is assessed on 11 benchmark datasets that are commonly used for CLIP scenario: Caltech101 \cite{fei2006one}, OxfordPets \cite{parkhi12cats}, StanfordCars \cite{krause20133object}, Flowers102 \cite{nilsback2008automated}, Food101 
\cite{bossard2014food}, FGVCAircraft \cite{maji2013fine}, SUN397 \cite{xiao2010sun}, DTD \cite{cimpoi14describing}, EuroSAT \cite{helber2019eurosat}, UCF101 \cite{soomro2012ucf101}, and ImageNet  \cite{ImageNet-IJCV-2015}. These datasets cover diverse classification challenges, from object recognition (Caltech101, Food101) to fine-grained tasks (StanfordCars, FGVCAircraft) and scene understanding (SUN397, EuroSAT). For each dataset, the vision-text feature vectors $\bs{z}_n$ are extracted using the CLIP's pre-trained model, while applying a temperature parameter $T_z$ as described in Appendix \ref{app:EM_Dir}. This temperature parameter has also been learned through the unrolling approach. 

\noindent\textbf{Results}: The proposed UNEM-Dirichlet has also been compared to its original EM-Dirichlet version \cite{EMDIRICHLET_CVPR2024}  as well as different few-shot classification methods. Table~\ref{tab:vlm_main_exp} depicts the accuracy results evaluated over 1,000 tasks with 4-shots. Thus, it can be noticed that UNEM-Dirichlet outperforms state-of-the-art methods for most datasets. In particular, unrolling the recent EM-Dirichlet algorithm yields further improvement of about 4.2\% in average. The achieved gain is more significant (reaching up to 7.5\%) with more challenging datasets having a large number of classes like FGVC Aircraft, Stanford Cars, and SUN397. This is due to the inappropriate choice of the hyperparameter $\lambda$ for these datasets. Indeed, the selected $\lambda$ parameter was set to $\frac{K}{K_{\rm eff}}|\query|$ in EM-Dirichlet approach  \cite{EMDIRICHLET_CVPR2024}. Thus, by computing the numerical values of the hyperparameter for each of the aforementioned datasets, it can be deduced from Figure \ref{fig:acc_lambda} that the difference between the selected value (which is 3,000 for Stanford Cars) and the optimal one (which is around 5,500 for Stanford Cars) is more important and impacts significantly the accuracy performance. This problem is well addressed by resorting to the proposed unrolled model for hyperparameters optimization. 

Appendix \ref{app:add_res} includes additional results  to illustrate, in both scenarios, the effects of the temperature parameter in this framework, and to show the benefits of learning variable hyperparameters across the architecture layers.

%% file: sec/5_conclusion.tex
\section{Conclusion}
\label{sec:conclusion}
In this paper, we focus on transductive few-shot methods grounded on a generalized form of the EM algorithm. These methods include a parameter to control class balance. The proposed GEM algorithm, applicable to any mixture of distributions, incorporates also a temperature scaling parameter. The optimization process is then unrolled into a neural network architecture, enabling efficient learning of the introduced hyper-parameters. The results demonstrate the effectiveness of the proposed approach on both vision-only and vision-language models. In future work, it would be valuable to explore the potential of unrolling techniques across a broader range of computer vision tasks, especially with the recent rise of foundational vision-language models. 

%% file: sec/X_suppl.tex
\clearpage
\setcounter{page}{1}
\maketitlesupplementary
\setcounter{subsection}{0}
\renewcommand\thesubsection{\Alph{subsection}}
\subsection{Details on the minimization steps of the GEM optimization algorithm}
\label{app:det_min_steps}
The optimization algorithm alternates between a minimization step
w.r.t. the distribution parameters and one w.r.t. the assignment variables.
In the following, $\ell$ designates the current iteration.\\
\noindent $\bullet$ \textbf{Minimization step w.r.t. the distribution parameter} \\
For every $k\in \{1,\ldots,K\}$,
the first estimation step w.r.t. $\bs{\theta}_k$, with $\bs{u}_n=(u_{n,k}^{(\ell)})_{1 \leq k \leq K}$ given, is performed by considering the following optimization problem:
\begin{align}
\underset{\bs{\theta}_k}{\mathrm{minimize}}\,   && - \sum_{n=1}^{N}u_{n,k}^{(\ell)} \ln \p \left( \bs{z}_{n} ~|~ {\bs{\theta}_k} \right ),
\end{align}
For a pdf belonging to the exponential family, this optimization problem is a convex. For instance, in the case of a Gaussian distribution whose pdf is defined in \eqref{eq:gauss_distribution}, the negative log-likelihood term, designated by function $F$, reduces to
\begin{align}
\label{gauss_log-likelihood}
F(\bs{\theta}_k)= \frac{1}{2} \sum_{n=1}^{N}u_{n,k}^{(\ell)} \|\bs{z}_n-\bs{\theta}_k \|^2.
\end{align}
The minimization of the above function \eqref{gauss_log-likelihood} w.r.t $\bs{\theta}_k$ results in an explicit form of the estimated distribution 
parameter $\bs{\theta}_k^{(\ell+1)}$ given by
\begin{align}
\label{eq:gaus_DP_est}
\bs{\theta}_k^{(\ell+1)}=\frac{\sum_{n=1}^{N}u_{n,k}^{(\ell)}\bs{z}_n}{\sum_{n=1}^{N}u_{n,k}^{(\ell)}}. 
\end{align}
In turn, in the case of Dirichlet distribution whose pdf is defined in \eqref{eq:dirichlet_distribution}, the negative log-likelihood term reads
\begin{multline}
\label{dir_log-likelihood}
F(\bs{\theta}_k)= \sum_{n = 1}^{N}  u_{n,k}^{(\ell)} \Bigg(-\sum_{i=1}^K (\theta_{k,i} -1) \ln z_{n,i} \\ + \sum_{i=1}^K \ln \Gamma(\theta_{k,i}) - \ln \Gamma\left(\sum_{i=1}^K \theta_{k,i}\right) \Bigg).
\end{multline}
Unlike the Gaussian model, the minimization of Dirichlet negative log-likelihood \eqref{dir_log-likelihood} has no closed form solution.
To circumvent this problem, we resort to the Majorization-Minorization (MM) strategy
recently developed in \cite{EMDIRICHLET_CVPR2024}. Thus, the estimated distribution parameter 
$\bs{\theta}_k^{(\ell+1)}$ can be expressed as follows
\begin{align}
\label{eq:DP_est_dir}
\bs{\theta}_k^{(\ell+1)}=\text{MM}(\bs{u}_{\cdot, k}^{(\ell)}, \bs{\theta}_k^{(\ell)}).
\end{align}  
\noindent $\bullet$ \textbf{Minimization step w.r.t. the assignment variable} \\
For every $n\in \query$, 
the second estimation step w.r.t. $\bs{u}_n$ is achieved by minimizing the objective function \eqref{eq:opt_prob}, 
while keeping the distribution parameter set to the estimated vector $\bs{\theta}_k^{(\ell+1)}$. However, since the partition complexity term 
$\Psi$ is non convex, it is replaced by a linear tangent upper bound. More specifically, the following tangent inequality can be used:
\begin{align}
\pi_k \ln \pi_k \ge \pi_k^{(\ell+1)} \ln \pi_k^{(\ell+1)} + (1+ \ln \pi_k^{(\ell+1)})(\pi_k-\pi_k^{(\ell+1)}) 
\end{align}
Knowing that $\pi_k= \frac{1}{|\query|} \sum_{n\in \query} u_{n,k}$, the optimization problem \eqref{eq:opt_prob} can be rewritten as follows
\begin{align}
\underset{\bs{u}_n}{\mathrm{minimize}} \quad G(\bs{u}_n)
\end{align}
with
\begin{multline}
\label{eq:obj_G}
G(\bs{u}_n) = - \sum_{k=1}^{K}u_{n,k} \ln \p \left( \bs{z}_{n} ~|~ {\bs{\theta}_k^{(\ell+1)}} \right ) \\
-\lambda 
\sum_{k=1}^{K} 
\frac{(1+\ln \pi_k^{(\ell+1)})}{|\query|}(u_{n,k}-u_{n,k}^{(\ell)}) 
\\
+T\sum_{k=1}^{K}u_{n,k}\ln u_{n,k} + \gamma_n \Bigg(\sum_{k=1}^{K}u_{n,k}-1\Bigg)
\end{multline} 
where $\gamma_n$ is a Lagrange multiplier aiming to enforce the sum-to-one constraint. The nonnegativity constraint can be dropped since we will show next that it is satisfied by the minimizer of $G$
subject to the sum-to-one constraint.
\\
The above optimization problem is convex.
By cancelling the derivative of the above objective function \eqref{eq:obj_G} w.r.t. $u_{n,k}$, it can 
be checked that
\begin{multline}
\label{eq:ln_u}
\ln u_{n,k}=-1-\frac{\gamma_n}{T} + \frac{1}{T}\Bigg( \ln \p \left( \bs{z}_{n} ~|~ {\bs{\theta}_k^{(\ell+1)}} \right ) \\ + \frac{\lambda}{|\query|}(1+\ln \pi_k^{(\ell+1)})\Bigg).
\end{multline}
By applying the exponential function to \eqref{eq:ln_u} and determining the multiplier $\gamma_n$ so that the sum-to-one constraint is satisfied, it can be deduced that the optimal class assignment vector $\bs{u}_n^{(\ell+1)}$ is obtained by applying the softmax function:
\begin{multline}
\label{eq:u_soft}
\bs{u}_n^{(\ell+1)}\\
=  \text{\normalfont softmax}\left( \frac{1}{T} \left(\ln \mathrm{p} \left( \boldsymbol{z}_{n} ~|~ {\bs{\theta}_k^{(\ell+1)}}  \right)  + \frac{\lambda}{|\query|}\ln(\pi_{k}^{(\ell+1)}) \right)_{k}  \right).
\end{multline}
\subsection{Generalized EM algorithm in the case of Gaussian distribution}\label{app:EM_Gaus}
\subsubsection{Feature representation in vision-only few-shot-setting} 
Let us consider a few-shot scenario for vision-only models. Thus, for all dataset samples $\bs{x}_n$ with $n\in \{1,\ldots,N\}$, the feature vectors $\bs{z}_n$ are generated using a visual feature extractor $f^{(v)}$ as follows
\begin{align}
\bs{z}_n=T_zf^{(v)}(\bs{x}_n)
\end{align}
where $T_z$ is a positive scaling parameter.\\


\subsubsection{Optimization algorithm}

Using \eqref{gauss_log-likelihood}, \eqref{eq:gaus_DP_est}, and \eqref{eq:u_soft},  the proposed GEM algorithm reduces to Algorithm \ref{algo:Gen_EM_G} in the case of a Gaussian distribution model.
\begin{algorithm}
\caption{GEM-Gaussian based few-shot classification algorithm}
\label{algo:Gen_EM_G}
\begin{algorithmic}
\STATE \textbf{Input:} Compute $\bs{z}_n$ for the dataset samples and, for all $k\in \{1, \dots, K\}$, initialize $\bs{\theta}_k^{(0)}$ as the means computed on the support set, and $\pi_k^{(0)}=1$ 
\FOR{$\ell = 0,1,\ldots,L-1$}
\STATE // \textcolor{gray}{Update assignment vectors for all query samples}
\STATE $\displaystyle \bs{u}_n^{(\ell +1)}$ \\ $=  \text{\normalfont softmax}\left( \frac{1}{T} \left(-\frac{1}{2}\|\bs{z}_n-\bs{\theta}_k^{(\ell)} \|^2+ \frac{\lambda}{|\query|}\ln(\pi_{k}^{(\ell)}) \right)_{ k}  \right) $ \\ 
\STATE // \textcolor{gray}{Update the mean parameter for each class}
\STATE $\displaystyle\bs{\theta}_k^{(\ell+1)} = \frac{\sum_{n=1}^{N}u_{n,k}^{(\ell+1)}\bs{z}_n}{\sum_{n=1}^{N}u_{n,k}^{(\ell+1)}}$,  \quad $\forall k \in\{1, \dots, K\},$\\
\STATE // \textcolor{gray}{Update class proportions}
\STATE $\displaystyle\pi_{k}^{(\ell+1)} = \frac{1}{|\query|} \sum_{n\in \query} u_{n, k}^{(\ell+1)}$,  \quad $\forall k \in\{1, \dots, K\},$
\ENDFOR
\end{algorithmic}
\end{algorithm}

\subsection{Generalized EM algorithm in the case of Dirichlet distribution}\label{app:EM_Dir}
\subsubsection{Feature representation in few-shot CLIP} 
Our second few-shot scenario is devoted to vision-language models such as CLIP. Let us assume \( f^{(v)} \) a vision-based feature extractor, and \( f^{(l)} \) a language-based feature extractor. Thus, for a sample \( \bs{x}_n \) with $n\in \{1,\ldots,N\}$ and a text prompt \( \bs{t}_k \) of class  \( k\in \{1,\ldots,K\} \) (for example $\bs{t}_k=$ \enquote{a \ photo \ of \ a \ $\{$class $k\}$}), the visual and text features are given by $f^{(v)}(\bs{x}_n)$ and $f^{(l)}(\bs{t}_k)$, respectively. Then, the resulting feature embeddings of the data sample $\bs{x}_n$ is defined as its probability vector of belonging to class $k$:

\begin{align}
\bs{z}_n=\text{softmax}\left\{ T_z \, \cos\left(f^{(v)}(\bs{x}_n), f^{(l)}(\bs{t}_k)\right)_{1 \leq k \leq K} \right\} ,
\end{align}
where $T_z > 0$ is a temperature scaling parameter.\\


\subsubsection{Optimization algorithm}
Using \eqref{eq:DP_est_dir} and \eqref{eq:u_soft}, and in the case of a Dirichlet data distribution model, the proposed GEM algorithm yields Algorithm \ref{algo:Gen_EM_D}.

\begin{algorithm}
\caption{GEM-Dirichlet based few-shot classification algorithm \label{algo:Gen_EM_D}}
\label{algo:Gen_EM_D}
\begin{algorithmic}
\STATE \textbf{Input:} Compute $\bs{z}_n$ for the dataset samples, initialize $\boldsymbol{u}_{n}^{(0)}=\bs{z}_n$, and $\boldsymbol{\theta}_{k}^{(0)}=\mathbf{1}_{K}$
\FOR{$\ell = 0,1,\ldots,L-1$} 
\STATE \textcolor{gray}{// Update the Dirichlet parameter for each class}
\STATE $\displaystyle\bs{\theta}_k^{(\ell +1)} = \text{MM}(\bs{u}_{\cdot, k}^{(\ell)}, \bs{\theta}_k^{(\ell)}) $,  \quad $\forall k \in\{1, \dots, K\},$\\
\STATE \textcolor{gray}{// Update class proportions}
\STATE $\displaystyle\pi_{k}^{(\ell+1)} = \frac{1}{|\query|} \sum_{n\in \query} u_{n, k}^{(\ell)}$,  \quad $\forall k \in\{1, \dots, K\},$\\
\STATE \textcolor{gray}{// Update assignment vectors for all query samples}
\STATE $\mathcal{L}_{n,k}^{(\ell)}= 
\sum_{i=1}^K (\theta_{k,i}^{(\ell+1)} -1) \ln z_{n,i}$\\ $\quad- \sum_{i=1}^K \ln \Gamma(\theta_{k,i}^{(\ell+1)}) + \ln \Gamma\left(\sum_{i=1}^K \theta_{k,i}^{(\ell+1)}\right)$
\STATE $\displaystyle \bs{u}_n^{(\ell +1)} $ \\ $=  \text{\normalfont softmax}\left( \frac{1}{T} \left(
\mathcal{L}_{n,k}^{(\ell)}
+ \frac{\lambda}{|\query|}\ln(\pi_{k}^{(\ell+1)}) \right)_{ k}  \right) $
\ENDFOR
\end{algorithmic}
\end{algorithm}

\subsection{Additional results}
\label{app:add_res}
\subsubsection{Ablation studies}
\label{app:ab_study}
In this part, we perform ablation studies to illustrate  the impact of the network depth, the effects of the introduced temperature scaling parameter and the benefits of learning adaptive hyper-parameters across the unrolled architecture layers. \\

\noindent$\bullet$ \textbf{Impact of the unrolled architecture depth}\\
First, we propose to analyze the impact of the number $L$ of layers of our unrolled architecture on the model accuracy, model size, and computational time. Table \ref{tab:imp_layers} reports the results. Thus, one of the main advantages of our UNEM model is that a few layers (about 7 or 10) are enough to achieve good performance. In what follows, the experiments are conducted using $L=10$. \\

\begin{table}[!h]
\centering
\scriptsize
    \begin{tabular}{l|c|c|c|c}
     \specialrule{0.8pt}{0pt}{1pt}
    \#Layers ($L$) & \#Params & Acc. & Train Time (s) &  Inference Time/task (s) \\
    3 &  7	& 65.6 & 2.80 &  $3.04 \times 10^-2$ \\
    5 &  11 & 65.9 & 3.22 &  $3.09 \times 10^-2$ \\
    7 &  15 & 66.3 & 3.46 &  $3.21 \times 10^-2$ \\
    10 & 21 & 66.4 & 3.61 &  $3.36 \times 10^-2$ \\
    12 & 25 & 66.2 & 4.06 &  $3.45 \times 10^-2$ \\
    15 & 31 & 66.1 & 4.29 &  $3.48 \times 10^-2$ \\
    18 & 37 & 66.2 & 5.74 &  $3.68 \times 10^-2$ \\
    \specialrule{0.8pt}{1pt}{0pt}
    \end{tabular}
    \vspace{-0.1cm}
    \caption{Impact of the number of layers ($L$) on UNEM-Gaussian performance using \textit{mini}-ImageNet, 5-shot and ResNet18.}
\label{tab:imp_layers}
\end{table}

\noindent $\bullet$ \textbf{Effects of temperature scaling} \\
To perform this study, we compare our unrolled architectures (UNEM-Gaussian as well as UNEM-Dirichlet) in both cases: (i) without introducing the temperature scaling parameter (as considered in the original algorithms PADDLE \cite{PADDLE_NEURIPS2022} and EM-Dirichlet \cite{EMDIRICHLET_CVPR2024}); (ii) while incorporating the temperature scaling (as proposed in our GEM algorithm). \\
Tables \ref{tab:suppl_t_gaussian_imagenet} and \ref{tab:suppl_t_cub_gaussian} depict the accuracy results in vision-only few-shot setting. Thus, it can be noticed that including the temperature scaling yields an accuracy improvement, which may vary from 1\% to 3\%. Moreover, in the context of vision-language models whose accuracy results are shown in Table \ref{tab:suppl_t_dirichlet}, similar gains (reaching up to 3\%), depending on the target donwstream dataset, are also achieved. This confirms again the advantage of incorporating the temperature scaling in our generalized algorithm. \\

\noindent $\bullet$ \textbf{Fixed vs adaptive hyper-parameters across layers}\\
One of the key advantages of unrolling algorithms is their flexibility in optimizing hyper-parameters, while allowing them to vary across the architecture layers. To show the potential of such hyper-parameter optimization approach, we propose to compare the proposed unrolled architectures (UNEM-Gaussian and UNEM-Dirichlet) in the following two cases: (i) the hyper-parameters are set fixed across the layers (as it is generally considered in original iterative algorithms), (ii) a set of hyper-parameters, adapted to the different layers, is learned. \\
Tables \ref{tab:suppl_hp_lay_gaussian_imagenet} and \ref{tab:suppl_hp_lay_cub_gaussian} provide the accuracy results for fixed and adaptive hyper-parameters optimization with vision-only models. It can be seen that learning adaptive hyper-parameters yields an accuracy gain of about 2-4\% compared to the case when the hyper-parameters are kept fixed across layers. Similar comparisons are also performed with vision-language models as shown in Table \ref{tab:suppl_hp_lay_vlm}. In this context, the improvement achieved by learning adaptive hyper-parameters often ranges from 1 to 2\%.  \\

\begin{table*}[!h]
    \centering
    \scalebox{0.90}
    {
    \begin{tabular}{l c ccc ccc}
        \toprule
        \multirow{2}{*}{\textbf{Temperature scaling}} & \multirow{2}{*}{\textbf{Backbone}} & \multicolumn{3}{c}{\textbf{\textit{mini}-ImageNet} (\textcolor{blue}{$K=20$})} & \multicolumn{3}{c}{\textbf{\textit{tiered}-ImageNet (\textcolor{blue}{$K=160$})}} \\
        \cmidrule(lr){3-5} \cmidrule(lr){6-8}
         & & 5-shot & 10-shot & 20-shot & 5-shot & 10-shot & 20-shot \\
        \bottomrule
          $\times$ & \multirow{2}{*}{ResNet-18} & 66.1 & 75.4	& 80.3 & 49.7 &	63.2 & 70.0 \\
       
         \checkmark & & \textbf{66.4} & \textbf{75.6} & \textbf{80.4} & \textbf{52.3} & \textbf{65.7} & \textbf{73.2}\\
        \midrule
        
         $\times$ & \multirow{2}{*}{WRN28-10} & \textbf{71.9} & 78.9 & 82.8 & 52.0 & 65.8 & 73.0 \\
    
        \checkmark & & 71.6 & \textbf{79.2} & \textbf{83.7} & \textbf{54.1} & \textbf{66.8} & \textbf{74.7} \\
        \bottomrule
    \end{tabular} 
    }
    \caption{Effects of the temperature scaling on the accuracy performance of UNEM-Gaussian approach applied to \textit{mini}-ImageNet and \textit{tiered}-ImageNet datasets.}
    \label{tab:suppl_t_gaussian_imagenet}
\end{table*}

\begin{table*}[!h]
    \centering
      \scalebox{0.90}
      {
    \begin{tabular}{l ccc}
        \toprule
        \multirow{2}{*}{\textbf{Temperature scaling}} & \multicolumn{3}{c}{\textbf{CUB} (\textcolor{blue}{$K=50$})} \\
        \cmidrule(lr){2-4} 
          & 5-shot & 10-shot & 20-shot \\
        \bottomrule
       
         $\times$ & 78.1 & 85.2	& 88.6 \\
        \checkmark & \textbf{78.5} & \textbf{85.3} & \textbf{88.6} \\ 
    \end{tabular} 
    }
    \caption{Effects of the temperature scaling on the accuracy performance of UNEM-Gaussian approach applied to CUB dataset.}
    \label{tab:suppl_t_cub_gaussian}
\end{table*}

\begin{table*}[!h]
    \centering
    \scalebox{0.94}{
    \begin{tabular}{c| cccccccccc}
        \textbf{Temperature scaling}
         & \rotatebox{60}{Food101} 
         & \rotatebox{60}{EuroSAT} 
         & \rotatebox{60}{DTD} 
         & \rotatebox{60}{OxfordPets} 
         & \rotatebox{60}{Flowers102} 
         & \rotatebox{60}{Caltech101} 
         & \rotatebox{60}{UCF101} 
         & \rotatebox{60}{FGVC Aircraft} 
         & \rotatebox{60}{Stanford Cars} 
         & \rotatebox{60}{SUN397}  \\

        \bottomrule

           $\times$ & 90.6 & 51.9 & \textbf{65.4} & 95.4 & 92.0 & 92.4 & \textbf{79.1} & 27.5 & 78.2 & 88.4   \\

           \checkmark & \textbf{91.4} & \textbf{53.8} & 65.3 & \textbf{96.0} & \textbf{95.6} & \textbf{93.4} & 78.5 & \textbf{30.4} & \textbf{80.0} & \textbf{88.5}  \\

    \end{tabular}
    }
    \caption{Effects of the temperature scaling on the accuracy performance of UNEM-Dirichlet approach applied to the vision-language models.}
    \label{tab:suppl_t_dirichlet}
\end{table*}
\begin{table*}[!h]
    \centering
    \scalebox{0.90}
    {
    \begin{tabular}{l c ccc ccc}
        \toprule
        \multirow{2}{*}{\textbf{Params across the layers}} & \multirow{2}{*}{\textbf{Backbone}} & \multicolumn{3}{c}{\textbf{\textit{mini}-ImageNet} (\textcolor{blue}{$K=20$})} & \multicolumn{3}{c}{\textbf{\textit{tiered}-ImageNet (\textcolor{blue}{$K=160$})}} \\
        \cmidrule(lr){3-5} \cmidrule(lr){6-8}
         & & 5-shot & 10-shot & 20-shot & 5-shot & 10-shot & 20-shot \\
        \bottomrule
          Fixed & \multirow{2}{*}{ResNet-18} & 62.5 & 72.5 & 78.0 & 49.8 & 63.6 & 70.4 \\
       
         Adaptive & & \textbf{66.4} & \textbf{75.6} & \textbf{80.4} & \textbf{52.3} & \textbf{65.7} & \textbf{73.2}\\
        \midrule
        
         Fixed & \multirow{2}{*}{WRN28-10} & 68.7 & 77.0 & 82.0 & 51.6 & 64.6 & 72.1 \\
    
        Adaptive & & \textbf{71.6} & \textbf{79.2} & \textbf{83.7} & \textbf{54.1} & \textbf{66.8} & \textbf{74.7} \\
        \bottomrule
    \end{tabular} 
    }
    \caption{Fixed vs adaptive hyper-parameters setting in the UNEM-Gaussian approach, using \textit{mini}-ImageNet and \textit{tiered}-ImageNet datasets.}
    \label{tab:suppl_hp_lay_gaussian_imagenet}
\end{table*}

\begin{table*}[!h]
    \centering
      \scalebox{0.90}
      {
    \begin{tabular}{l ccc}
        \toprule
        \multirow{2}{*}{\textbf{Params across layers}} & \multicolumn{3}{c}{\textbf{CUB} (\textcolor{blue}{$K=50$})} \\
        \cmidrule(lr){2-4} 
          & 5-shot & 10-shot & 20-shot \\
        \bottomrule
       
         Fixed & 75.2 & 82.9 & 87.1	 \\
         Adaptive & \textbf{78.5} & \textbf{85.3} & \textbf{88.6} \\ 
    \end{tabular} 
    }
    \caption{Fixed vs adaptive hyper-parameters setting in the UNEM-Gaussian approach, using CUB dataset.}
    \label{tab:suppl_hp_lay_cub_gaussian}
\end{table*}

\begin{table*}[!h]
    \centering
    \scalebox{0.94}{
    \begin{tabular}{c| cccccccccc}
        \textbf{Params across layers}
         & \rotatebox{60}{Food101} 
         & \rotatebox{60}{EuroSAT} 
         & \rotatebox{60}{DTD} 
         & \rotatebox{60}{OxfordPets} 
         & \rotatebox{60}{Flowers102} 
         & \rotatebox{60}{Caltech101} 
         & \rotatebox{60}{UCF101} 
         & \rotatebox{60}{FGVC Aircraft} 
         & \rotatebox{60}{Stanford Cars} 
         & \rotatebox{60}{SUN397}  \\

        \bottomrule

           Fixed & 89.6 & 52.2 & 64.8 & 95.3 & 95.3 & 92.3 & \textbf{79.2} & \textbf{31.6} & 78.0 & 87.6 \\

           Adaptive & \textbf{91.4} & \textbf{53.8} & \textbf{65.3} & \textbf{96.0} & \textbf{95.6} & \textbf{93.4} & 78.5 & 30.4 & \textbf{80.0} & \textbf{88.5} \\

    \end{tabular}
    }
    \caption{Fixed vs adaptive hyper-parameters setting in the UNEM-Dirichlet approach, using the vision-language models.}
    \label{tab:suppl_hp_lay_vlm}
\end{table*}

\subsubsection{Performance under distribution shifts}
In Table \ref{tab:dist-shift}, we include the accuracy of the original PADDLE algorithm \cite{PADDLE_NEURIPS2022} and its unrolled version, with \textit{tiered}-ImageNet used for pre-training, and a fine-grained classification dataset (CUB)  as well as \textit{mini}-ImageNet used for inference. One could observe similar improvements (about 4-7\%) brought by UNEM.
\begin{table*}[!h]
\scriptsize
\centering
 \begin{tabular}{l|c|c}
 \specialrule{0.8pt}{0pt}{1pt}
 Method & CUB & \textit{mini}-ImageNet\\
 PADDLE \cite{PADDLE_NEURIPS2022} & 66.0 & 82.9 \\
 \textbf{UNEM-Gaussian} & \textbf{72.9} & \textbf{87.0} \\
 \specialrule{0.8pt}{1pt}{0pt}
 \end{tabular}
 \caption{Cross-domain evaluation: Accuracy performance on \textit{mini}-ImageNet and CUB using a model trained on \textit{tiered}-ImageNet, with a 5-shot setting and a ResNet18 backbone.}
\label{tab:dist-shift}
\end{table*}

\subsubsection{Backbone effect in CLIP}
The performance of EM-Dirichlet and UNEM-Dirichlet using ViT-B/32 as backbone are shown in Table \ref{tab:backbone}. Thus, in comparison to CLIP with ResNet50 (see Table \ref{tab:vlm_main_exp}), one may observe a similar or slightly better accuracy performance with most datasets. Moreover, the gains brought by UNEM over the iterative variant are consistent with those observed with ResNet50 backbone.
\begin{table*}[!h]
\centering
 \begin{tabular}{l| cccccccccc}
    Method 
    & \rotatebox{60}{Food101} 
    & \rotatebox{60}{EuroSAT} 
    & \rotatebox{60}{DTD} 
    & \rotatebox{60}{OxfordPets} 
    & \rotatebox{60}{Flowers102} 
    & \rotatebox{60}{Caltech101} 
    & \rotatebox{60}{UCF101} 
    & \rotatebox{60}{FGVC Aircraft} 
    & \rotatebox{60}{Stanford Cars} 
    & \rotatebox{60}{SUN397}  \\

    \bottomrule
      
 EM-Dirichlet \cite{EMDIRICHLET_CVPR2024} & 89.3 & 54.8 & 63.9 & 92.5 & 92.7 & 92.8 & 77.2 & 27.3 & 73.9 & 81.7\\
 \textbf{UNEM-Dirichlet} & \textbf{91.4} & \textbf{57.5} & \textbf{67.4} & \textbf{95.7} & \textbf{95.2} & \textbf{94.0} & \textbf{80.4} & \textbf{33.8} & \textbf{77.8} & \textbf{88.4} \\
 \end{tabular}
 \caption{Accuracy performance of iterative EM-Dirichlet \cite{EMDIRICHLET_CVPR2024} and our UNEM variant using CLIP ViT-B/32 for feature extraction.}
\label{tab:backbone}
\end{table*}
\begin{figure*}[!h]
    \centering
    \includegraphics[width=0.95\textwidth]{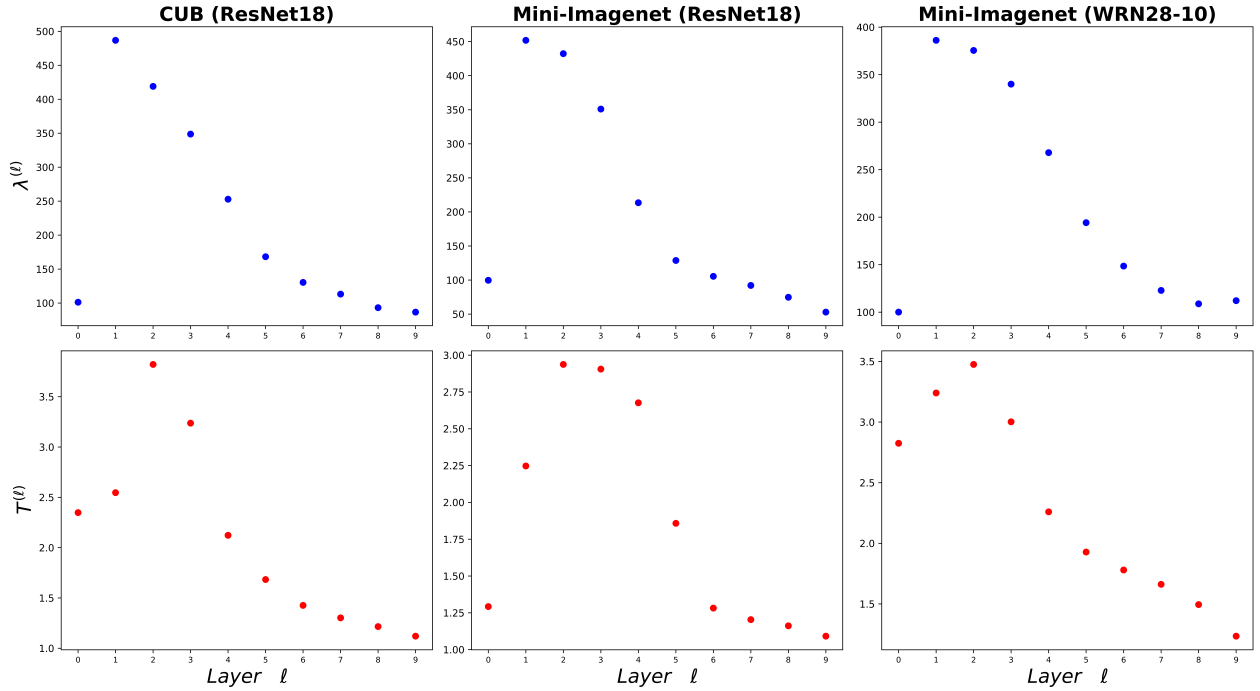} 
    \caption{Illustration of the learned hyper-parameters $\lambda^{(\ell)}$ and $T^{(\ell)}$ across layers for CUB (with ResNet18 model), \textit{mini}-ImageNet (with ResNet18 model) and \textit{mini}-ImageNet (with WRN28-10 model).}
    \label{fig:Hyp_param_across_lay_vision_only}
\end{figure*}
\begin{figure*}[!h]
    \centering
    \includegraphics[width=0.95\textwidth]{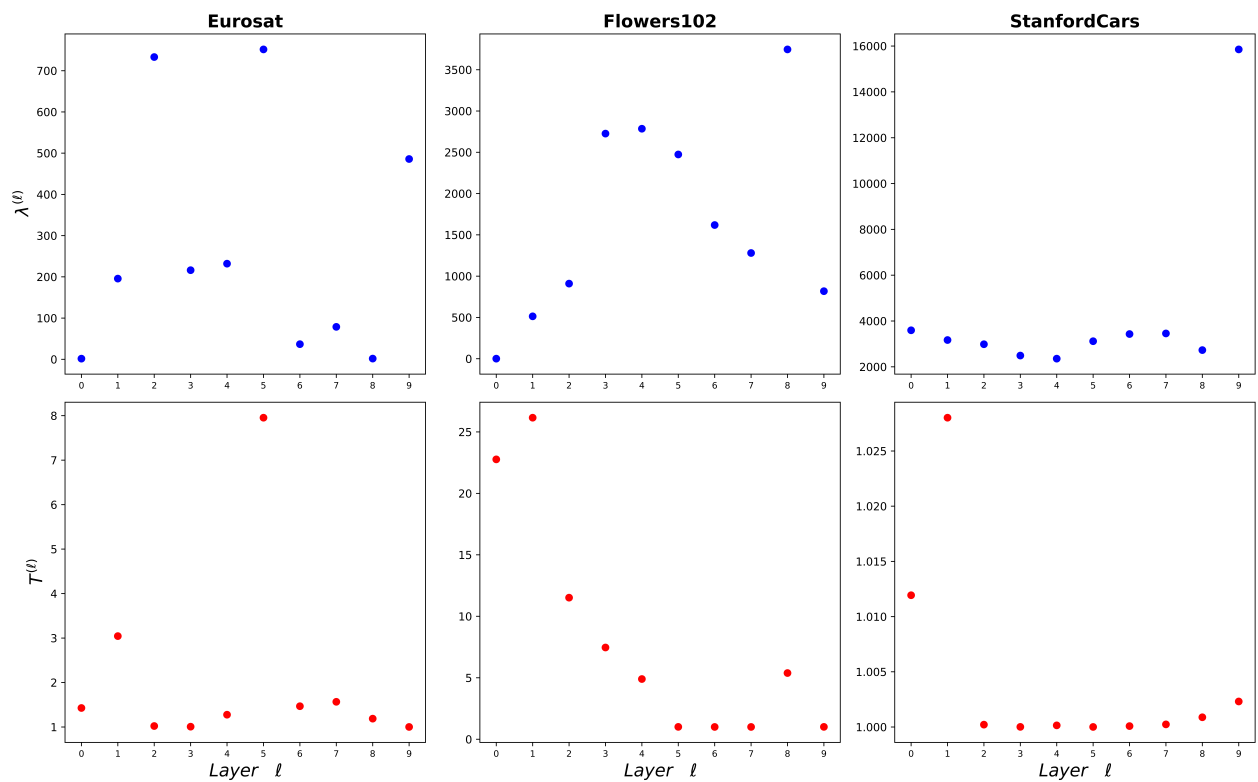} 
    \caption{Illustration of the learned hyper-parameters $\lambda^{(\ell)}$ and $T^{(\ell)}$ across layers for some datasets with vision-language models.}
    \label{fig:Hyp_param_across_lay_VLM}
\end{figure*}

\subsection{Illustration and analysis of the learned hyperparameters}
\label{app:add_res_hyper}
In this part, we propose to illustrate the variations of the learned hyper-parameters and analyze their orders-of-magnitude. \\

\noindent $\bullet$ \textbf{Illustration of the learned hyper-parameters} \\ 
The evolutions of the learned hyper-parameters $\lambda^{(\ell)}$ and $T^{(\ell)}$ with respect to the layer index are illustrated in Figures \ref{fig:Hyp_param_across_lay_vision_only} and \ref{fig:Hyp_param_across_lay_VLM} for some downstream image classification tasks. While 
Figure \ref{fig:Hyp_param_across_lay_vision_only} shows that the learned hyper-parameters with CUB (ResNet18), \textit{mini}-ImageNet (ResNet18), and \textit{mini}-ImageNet (WRN28-10) have similar amplitudes, much different orders-of-magnitude are observed with vision-language models as shown in Figure~\ref{fig:Hyp_param_across_lay_VLM} for some test datasets. Let us recall that the different learned hyper-parameters, for all datasets, are available at \url{https://github.com/ZhouLong0/UNEM-Transductive}. \\

\noindent $\bullet$ \textbf{Analysis of the learned hyper-parameters} \\
Different observations could be made from the previous illustrations. On the one hand, in the case of vision-only models, it can be seen that the learned hyper-parameters $\lambda^{(\ell)}$ appear quite similar. However, the evolution of $T^{(\ell)}$ values shows different behaviors.  
Moreover, it is important to note that the optimal hyper-parameters also depend on the pre-training model as observed with \textit{mini}-ImageNet (ResNet18) and \textit{mini}-ImageNet (WRN28-10). On the other hand, with vision-language models, it can be observed that both hyper-parameter values $\lambda^{(\ell)}$ and $T^{(\ell)}$ strongly depend on the target dataset. Indeed, unlike the vision-only models where the feature vectors have a fixed size (which is equal to the dimension of the pre-trained model's output), the feature vectors $\bs{z}_n$ in the context of few-shot CLIP have different sizes, depending on the number of classes of each target dataset. For instance, knowing that  EuroSAT, Flowers102 and Stanford Cars have 10, 102, and 196 classes, respectively; it can be observed that the smallest (resp. largest) values of $\lambda^{(\ell)}$ are obtained with EuroSAT (resp. Stanford Cars). These results are expected since, by increasing the dimension of $\bs{z}_n$, the magnitude of the log-likelihood term may increase, and so, a higher value of $ \lambda^{(\ell)}$ is needed to mitigate the class-balance bias. \\
This study shows the dependence of the introduced hyper-parameters on the target downstream dataset as well as the pre-training model, and confirms the importance of optimizing hyper-parameters in both evaluation scenarios.